\documentclass[runningheads]{llncs}
\usepackage[T1]{fontenc}
\usepackage{graphicx}
\usepackage{amsmath}
\usepackage{amssymb}
\usepackage{amsfonts}
\usepackage{bm}
\usepackage{booktabs}
\usepackage{multirow}
\usepackage{color}
\usepackage{url}
\usepackage{marvosym}

\begin{document}
\title{VoxShield: Protecting 3D Medical Datasets from Unauthorized Training via Frequency-Aware Inter-Slice Disruption}
\titlerunning{VoxShield}
\author{
Xinyao Liu\inst{1,2} \and
Zhipeng Deng\inst{1}\textsuperscript{\Letter} \and
Wenhan Jiang\inst{1} \and
Haolin Wang\inst{3} \and
Xun Lin\inst{4} \and
Yafei Ou\inst{5} \and
Yefeng Zheng\inst{1}\textsuperscript{\Letter}
}
\authorrunning{X. Liu et al.}
\institute{
Westlake University, Hangzhou, China
\and
Dalian University of Technology, Dalian, China
\and
Hokkaido University, Sapporo, Japan
\and
The Chinese University of Hong Kong, Hong Kong SAR, China
\and
RIKEN, Japan
\\
\email{
xinyao.liu266@outlook.com; \{dengzhipeng, jiangwenhan, zhengyefeng\}@westlake.edu.cn;
}
}

\maketitle
%
\begin{abstract}
The release of public 3D medical image segmentation (MIS) datasets accelerates clinical research but simultaneously heightens risks of unauthorized AI model training. While \textbf{Unlearnable Examples (UE)} offer protection by injecting imperceptible perturbations to prevent effective model learning, existing methods primarily target 2D scenarios. They neglect the \textit{volumetric spatial correlations} and \textit{inter-slice anatomical consistency} inherent in 3D medical volumes, which serve as critical learning priors for 3D segmentation networks. To bridge this gap, we propose \textbf{VoxShield}, a UE framework that explicitly targets the volumetric inductive biases of 3D networks. Our core insight is that by systematically dismantling the cross-slice continuity that 3D architectures rely on, we can fundamentally impair their spatial aggregation process. Specifically, we introduce an \textit{Inter-Slice Frequency Consistency Disruption} mechanism that maximizes the spectral divergence between adjacent slices, injecting structural incoherence along the $z$-axis. 
Complementing this structural attack, a \textit{Semantic Prediction Disruption} module is incorporated. By maximizing the $\ell_1$ divergence between clean and perturbed logits, it forces the injected noise to penetrate the entire network and corrupt the final semantic mapping. Experiments on BraTS19 and FLARE21 demonstrate that VoxShield successfully degrades 3D segmentation performance, reducing the DSC from 80.0\% to near 0.0\% and from 88.6\% to 6.8\%, respectively. All protections are achieved with minimal perturbation ($\epsilon{=}4/255$) to preserve high visual fidelity. The code is available at \url{https://github.com/KK266299/VoxShield}.

\end{abstract}
%
\section{Introduction}
\begin{figure}[t]
\centering
\includegraphics[width=\linewidth]{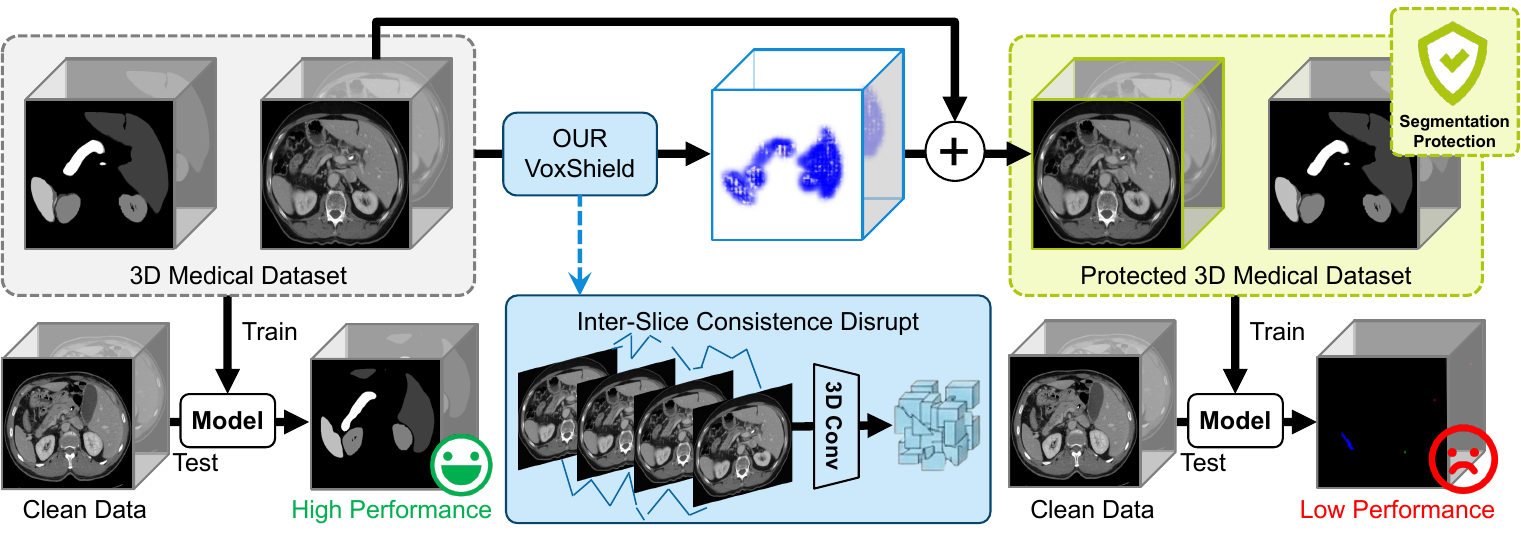}
\caption{The defense mechanism of VoxShield. \textbf{Left:} A model trained on clean data learns coherent volumetric representations and segments accurately on clean test volumes. \textbf{Right:} VoxShield disrupts inter-slice consistency by injecting spectrally diverse perturbations across adjacent slices. Consequently, a model trained on such protected data fails to extract coherent cross-slice features, yielding severely degraded segmentation on clean test inputs.}
\label{fig:motivation}
\end{figure}

3D medical image segmentation (MIS) is fundamental to disease diagnosis, treatment planning, and surgical navigation. The success of modern 3D segmentation models~\cite{cciccek20163d,milletari2016v,isensee2021nnu,hatamizadeh2022unetr,tang2022swinunetr} critically depends on large-scale, voxel-annotated datasets. To facilitate clinical collaboration and education, institutions increasingly share these valuable datasets through public repositories like TCIA~\cite{tcia}, often with complementary findings published in literature indexed by PubMed~\cite{pubmed}. Crucially, such sharing is typically governed by strict data-use agreements and licensing terms (e.g., CC BY-NC) that permit educational and non-commercial research, while restricting unauthorized AI model training.
Unfortunately, legal restrictions alone cannot physically prevent data scraping. Once in the public domain, 3D medical datasets are routinely downloaded and repurposed to train unauthorized networks. This misuse violates intellectual property rights and privacy laws, including the GDPR~\cite{gdpr2016,kaissis2020end}, highlighting an urgent need for proactive, technical data protection mechanisms.

To counter such risks, Unlearnable Examples (UE)~\cite{huang2021unlearnable} offer an appealing proactive defense by injecting imperceptible perturbations into training data \textit{before} release. Models trained on the perturbed data learn ``shortcut'' patterns instead of genuine semantic features, and consequently fail to generalize to clean test inputs. Representative methods include EM~\cite{huang2021unlearnable}, which crafts error-minimizing noise via bi-level optimization; TAP~\cite{fowl2021adversarial}, which leverages targeted adversarial perturbations; SEP~\cite{liu2023stable}, which stabilizes noise through a stable error-minimizing objective; LSP~\cite{yu2022availability}, which aligns perturbations with linearly separable features; and CUDA~\cite{sadasivan2023cuda}, which generates class-wise perturbations via convolution. While UE have been explored for 2D image classification and more recently extended to 2D segmentation~\cite{sun2024unseg} and 3D point clouds~\cite{wang2024unlearnable3d}, existing methods remain ineffective for 3D medical volumetric segmentation due to a critical architectural blind spot: they neglect the volumetric priors unique to 3D data.

Modern 3D segmentation architectures rely on a key inductive bias: \textit{anatomical inter-slice consistency}~\cite{isensee2021nnu,dong2022mnet}. Organs and lesions vary smoothly along the $z$-axis, and 3D convolution kernels explicitly aggregate cross-slice context to exploit this continuity. The importance of this prior is evidenced by a well-established empirical finding: under severe anisotropy, where large inter-slice spacing naturally disrupts volumetric continuity, 2D models consistently outperform their 3D counterparts~\cite{isensee2021nnu,dong2022mnet}. Although voxel spacing is an intrinsic acquisition parameter that cannot be altered post hoc, this observation reveals that inter-slice consistency is the key factor governing 3D segmentation performance. This further suggests that deliberately disrupting the cross-slice aggregation process of 3D convolutions via injecting inter-slice incoherent perturbations along the $z$-axis can directly undermine the very inductive bias these models depend on, thereby providing a principled attack vector against 3D architectures.

As illustrated in Fig.~\ref{fig:motivation}, our key insight is that inter-slice consistency is simultaneously the source of 3D segmentation power and its exploitable vulnerability. Adjacent slices of a volume naturally share similar in-plane spatial patterns, which is a property that 3D convolutions heavily rely on to aggregate contextual cues along~$z$. We exploit this by injecting perturbations whose \emph{in-plane spectral characteristics} vary maximally between neighboring slices, so that 3D kernels spanning the $z$-axis receive conflicting frequency patterns at every spatial position, fundamentally undermining their ability to learn coherent volumetric representations.

Building on this insight, we propose \textbf{VoxShield}, a UE framework
tailored for 3D medical image segmentation. VoxShield comprises two core
modules: (i)~an \textit{Inter-Slice Frequency Consistency Disruption}
module that maximizes spectral discrepancies between adjacent slices in
the frequency domain, injecting ``inter-slice flickering'' that directly
undermines the $z$-axis continuity assumption of 3D models; and (ii)~a
\textit{Semantic Prediction Disruption} module that maximizes the
$\ell_1$ distance between clean and perturbed logits, forcing
perturbations to propagate through the entire network and corrupt its
semantic understanding.

\section{Methodology}

\subsection{Preliminaries}
\label{sec:preliminary}

\textbf{3D Medical Image Segmentation.}\quad
Given a clean training set
$\mathcal{D}_{\mathrm{clean}}\!=\!\{(\bm{x}_i,\bm{y}_i)\}_{i=1}^{N}$,
where $\bm{x}_i\!\in\!\mathbb{R}^{C\times D\times H\times W}$ is a 3D medical volume with $D$ slices and
$\bm{y}_i\!\in\!\{0,1\}^{D\times H\times W\times K}$ is the
corresponding voxel-wise annotation over $K$ classes,
3D medical image segmentation aims to optimize a network
$\mathcal{F}(\cdot\,;\theta)$ to establish the volumetric mapping from
$\bm{x}$ to $\bm{y}$.
This optimization can be formulated as:
\begin{equation}
  \arg\min_{\theta}\;
  \mathbb{E}_{(\bm{x},\bm{y})\sim\mathcal{D}_{\mathrm{clean}}}
  \bigl[\mathcal{L}_{\mathrm{seg}}\!\bigl(
    \mathcal{F}(\bm{x};\theta),\,\bm{y}\bigr)\bigr],
  \label{eq:seg}
\end{equation}
where $\mathcal{L}_{\mathrm{seg}}$ is the segmentation loss
(\textit{e.g.}, cross-entropy loss and Dice loss)
and $\theta$ denotes the trainable parameters of $\mathcal{F}$.

\noindent\textbf{Unlearnable Examples.}\quad
UE are designed to prevent unauthorized models from extracting useful
knowledge by adding imperceptible perturbations to the training
images of $\mathcal{D}_{\mathrm{clean}}$.
For better understanding, we take the representative error-minimizing
(EM) noise~\cite{huang2021unlearnable} as an example.
EM jointly optimizes a surrogate model and sample-wise perturbations
to generate protective noise:
\begin{equation}
  \min_{\theta}\;
  \mathbb{E}_{(\bm{x},\bm{y})\sim\mathcal{D}_{\mathrm{clean}}}
  \Bigl[\min_{\bm{\delta}\in\mathcal{I}}
  \mathcal{L}_{\mathrm{seg}}\!\bigl(
    \mathcal{F}_s(\bm{x}+\bm{\delta};\theta),\,\bm{y}
  \bigr)\Bigr],
  \label{eq:ue}
\end{equation}
where $\mathcal{F}_s$ is a surrogate model,
$\mathcal{I}\!=\!\{\bm{\delta}:\|\bm{\delta}\|_\infty\!\le\!\epsilon\}$
is the feasible region that ensures imperceptibility, and
$\bm{x}+\bm{\delta}$ denotes the protected image (also called the
unlearnable example).
In this process, $\theta$ and $\bm{\delta}$ are alternately updated
to minimize $\mathcal{L}_{\mathrm{seg}}$, yielding the protected
dataset
$\mathcal{D}_{\mathrm{protect}}\!=\!\{(\bm{x}_i+\bm{\delta}_i,\,
\bm{y}_i)\}_{i=1}^{N}$.

\noindent\textbf{Task Objective.}\quad
The \textit{data protector} has access to
$\mathcal{D}_{\mathrm{clean}}$ and publishes
$\mathcal{D}_{\mathrm{protect}}$. The \textit{data exploiter} trains
a 3D segmentation architecture on
$\mathcal{D}_{\mathrm{protect}}$.
We evaluate the protection effectiveness by testing the model trained
by the exploiter on held-out clean data: if the test-time performance
degrades significantly, the protection is considered successful.
The protector aims to minimize such test-time performance while
keeping perturbations imperceptible.

\subsection{VoxShield Framework}
\label{sec:VoxShield}
\begin{figure}[t]
  \centering
  \includegraphics[width=\textwidth]{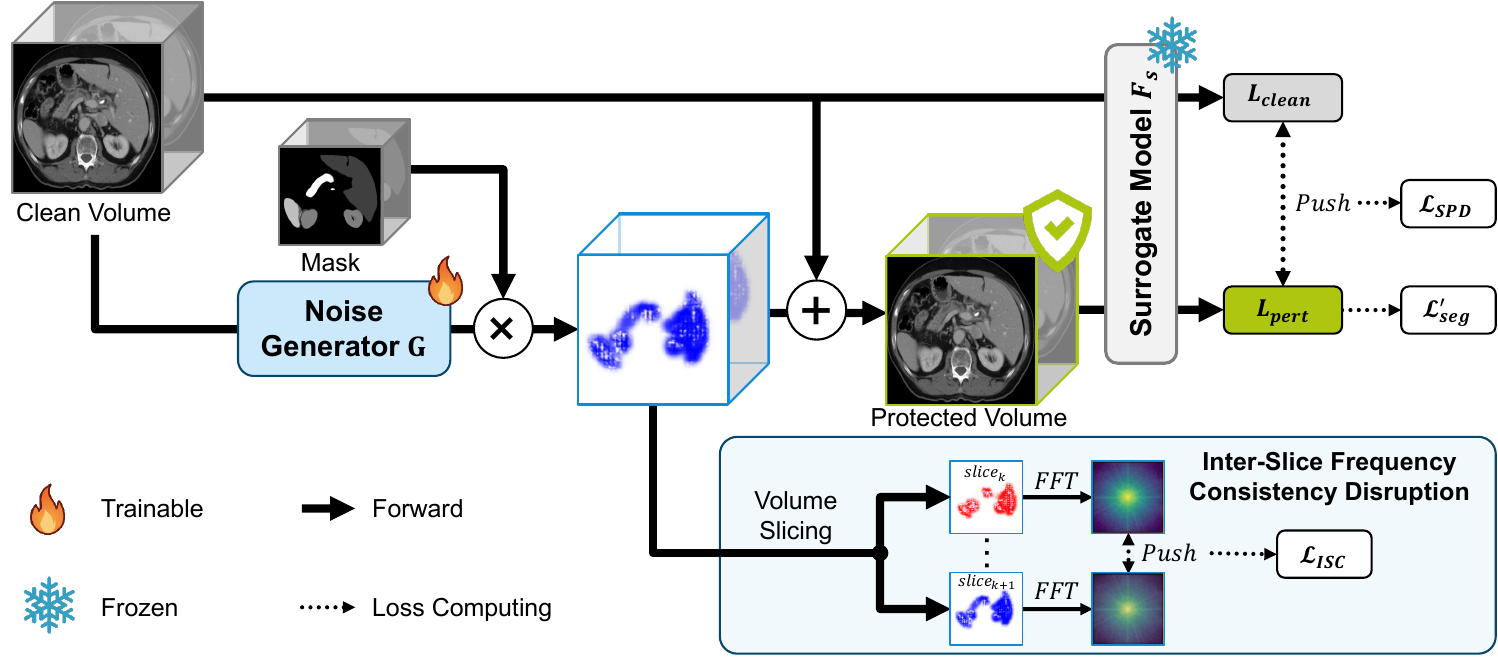}
\caption{Overview of the proposed VoxShield framework. A noise
generator $\mathcal{G}$ produces perturbations that are added to clean
3D medical images to create protected volumes. A surrogate model
$F_s$ computes $\mathcal{L}_{\text{SPD}}$ and $\mathcal{L}'_{\text{seg}}$ to drive protected images away from the original decision boundaries.
An inter-slice frequency consistency disruption module enforces
$\mathcal{L}_{\text{ISC}}$ by maximizing spectral discrepancy between
adjacent slices, destroying the volumetric continuity that 3D models
rely on.}
  \label{fig:method}
\end{figure}

\textbf{Overview.}\quad
Existing UE fail in 3D medical segmentation because they ignore
\textit{inter-slice anatomical consistency}, the key inductive bias
exploited by 3D models. As illustrated in Fig.~\ref{fig:method},
VoxShield introduces two dedicated modules: (i)~\textit{Inter-Slice Frequency Consistency Disruption} maximizes the spectral
discrepancy of perturbations between adjacent slices in the frequency
domain, breaking the inter-slice continuity prior that 3D models rely on;
(ii)~\textit{Semantic Prediction Disruption} maximizes the $\ell_1$
distance between clean and perturbed logits to corrupt semantic
predictions.
To focus perturbations on anatomically meaningful regions, the raw
generator output is masked by a binary region-of-interest mask
$\bm{M}\!\in\!\{0,1\}^{D\times H\times W}$ derived from $\bm{y}$:
\begin{equation}
  \bm{\delta} = \operatorname{clamp}\!\bigl(
G(\bm{x})\odot\bm{M},\;-\epsilon,\;\epsilon\bigr),
  \label{eq:mask}
\end{equation}
where $\odot$ denotes element-wise multiplication.
The protected volume is then $\bm{x}_p\!=\!\bm{x}+\bm{\delta}$.

\subsection{Inter-Slice Frequency Consistency Disruption}
3D segmentation networks fundamentally assume \textit{volumetric
continuity}, i.e., anatomical structures change smoothly along $z$.
3D convolution kernels with receptive fields spanning $k_z$ slices
aggregate cross-slice features under this assumption.
Our strategy is to violate this assumption by forcing the
perturbation's spectral characteristics to vary \textit{maximally}
between consecutive slices.

For each slice $z$, we compute the 2D amplitude
spectrum of the perturbation:
$\bm{S}_z=|\mathcal{F}_{2\text{D}}(\bm{\delta}_z)|\in\mathbb{R}^{H\times W}$,
where $\bm{\delta}_z\!\in\!\mathbb{R}^{H\times W}$ is the
$z$-th slice of the generated noise. The ISC loss is:
\begin{equation}
  \mathcal{L}_{\text{ISC}}
  =-\frac{1}{D{-}1}\sum_{z=1}^{D-1}
    \bigl\|\bm{S}_{z+1}-\bm{S}_z\bigr\|_2.
  \label{eq:isc}
\end{equation}
Minimizing $\mathcal{L}_{\text{ISC}}$ (i.e., maximizing inter-slice spectral
divergence) forces drastic spectral shifts between adjacent slices,
injecting high-frequency ``spectral jitter'' along $z$ that
fundamentally contradicts the smooth anatomical continuity 3D
convolutions rely on.
We measure divergence in the \textit{frequency domain} rather than
the spatial domain because magnitude spectra are
translation-invariant~\cite{yin2019fourier,wang2020high}, providing a more stable
and texture-focused measure of inter-slice dissimilarity.
\subsection{Semantic Prediction Disruption}
Effective data protection requires that perturbations corrupt not only
low-level pixel statistics but also the high-level semantic mapping
learned by the surrogate. If the surrogate can still produce similar
predictions on clean and perturbed inputs, the learned representations
remain intact and protection fails. We therefore explicitly maximize
the discrepancy between the surrogate's outputs on clean and perturbed
volumes, forcing perturbations to propagate through the entire network
and disrupt its semantic understanding.

Let $\bm{L}_{\text{clean}}\!=\!\mathcal{F}_s(\bm{x};\theta)$ and $\bm{L}_{\text{pert}}\!=\!\mathcal{F}_s(\bm{x}_p;\theta)$ denote the clean and perturbed logit volumes ($\in\mathbb{R}^{D\times H\times W\times K}$), respectively.
We maximize the $\ell_1$ distance between them:
\begin{equation}
  \mathcal{L}_{\text{SPD}} = -\|\,\bm{L}_{\text{pert}} - \bm{L}_{\text{clean}}\|_1.
  \label{eq:sp}
\end{equation}

Compared with the $\ell_2$ norm, the $\ell_1$ norm is less dominated by a few large-deviation voxels, thereby encouraging the generator to degrade predictions broadly across the volume rather than at isolated locations.

\subsection{Optimization}
The total generator objective combines segmentation, logits
divergence, and inter-slice consistency disruption:
\begin{equation}
  \mathcal{L}_{\mathrm{total}}
  =\mathcal{L}'_{\mathrm{seg}}\!\big(\mathcal{F}_s(\bm{x}_p;\theta),\,\bm{y}\big)
  +\lambda_{\text{SPD}}\,\mathcal{L}_{\text{SPD}}
  +\lambda_{\text{ISC}}\,\mathcal{L}_{\text{ISC}},
  \label{eq:total}
\end{equation}
where $\mathcal{L}'_{\text{seg}}$ (the combination of Dice and cross entropy loss) anchors training stability by ensuring the surrogate continues to learn on perturbed data (thus providing meaningful gradients), while $\mathcal{L}_{\text{SPD}}$ and $\mathcal{L}_{\text{ISC}}$ inject semantic and structural corruption, respectively. $\lambda_{\text{SPD}}$ and $\lambda_{\text{ISC}}$ balance the three terms.

The optimization of VoxShield follows a bi-level alternating strategy between the surrogate model $\mathcal{F}_s$ and the noise generator $\mathcal{G}$. During the generator's update phase, $\mathcal{F}_s$ is frozen to provide consistent adversarial gradients for perturbation refinement. This iterative cycle ensures that $\mathcal{G}$ learns to synthesize noise patterns that are specifically destructive to 3D volumetric priors. The detailed optimization steps for the generator are summarized in Algorithm~1. Upon convergence, the finalized generator is deployed as a fixed protective operator to transform the clean dataset $\mathcal{D}_{\mathrm{clean}}$ into its unlearnable counterpart, $\mathcal{D}_{\mathrm{protect}}$.

\begin{table}[t]
\centering
\small
\setlength{\tabcolsep}{3pt}
\begin{tabular}{p{0.95\linewidth}}
\toprule
\textbf{Algorithm 1:} VoxShield Training Procedure \\
\midrule
\textbf{Input:} Clean dataset $\mathcal{D}_{\mathrm{clean}}$,
  surrogate $\mathcal{F}_s$, generator $\mathcal{G}$, budget
  $\epsilon$, weights $\lambda_{\text{SPD}},\lambda_{\text{ISC}}$, learning rates $\alpha_g$ \\
\textbf{Output:} Protected dataset $\mathcal{D}_{\mathrm{protect}}$ \\[2pt]
1: \textbf{for} epoch $= 1$ \textbf{to} $n$ \textbf{do} \\
2: \quad \textbf{for} each $(\bm{x},\bm{y})$ in
  $\mathcal{D}_{\mathrm{clean}}$ \textbf{do} \\
3: \quad\quad $\bm{\delta}\leftarrow \mathrm{Clamp}_{[-\epsilon,\epsilon]}[\mathcal{G}(\bm{x})\odot \bm{M}];
~~\bm{x}_p\leftarrow\mathrm{Clamp}_{[0,1]}[\bm{x}+\bm{\delta}]$ \\
4: \quad\quad Compute $\mathcal{L}_{\mathrm{total}}$
  \hfill \textit{// Eq.~\eqref{eq:total}} \\
5: \quad\quad $\theta_{\mathcal{G}}\leftarrow
  \theta_{\mathcal{G}}-\alpha_g\nabla_{\theta_{\mathcal{G}}}
  \mathcal{L}_{\mathrm{total}}$ \\
6: \quad \textbf{end for} \\
7: \textbf{end for} \\
8: \textbf{for} each $(\bm{x},\bm{y})$ in
  $\mathcal{D}_{\mathrm{clean}}$ \textbf{do} \\
9: \quad $\mathcal{D}_{\mathrm{protect}}\leftarrow
  \mathcal{D}_{\mathrm{protect}}\cup
  \{(\mathrm{Clamp}_{[0,1]}[\bm{x}+\mathcal{G}(\bm{x})],
  \;\bm{y})\}$ \\
10: \textbf{end for} \\
11: \textbf{return} $\mathcal{D}_{\mathrm{protect}}$ \\
\bottomrule
\end{tabular}
\label{alg:VoxShield}
\end{table}

\section{Experiments and Results}

\subsection{Experimental Setup}

\noindent\textbf{Datasets.}
We evaluated the protection effectiveness of VoxShield on two 3D MIS datasets:
\textbf{BraTS19}~\cite{menze2014multimodal} (brain tumor MRI) and
\textbf{FLARE21}~\cite{ma2022flare} (abdominal organ CT), covering different anatomical regions and imaging modalities. All volumes were resized to $160 \times 160 \times 160$, and intensities were normalized to the range of $[0, 1]$.

\noindent\textbf{Models.}
We used a 3D UNet~\cite{cciccek20163d} implemented in MONAI for both the surrogate model and the generator for UE generation. To evaluate cross-architecture transferability, we trained four
victim models: ~\textit{3D-UNet}~\cite{cciccek20163d}, ~\textit{Attention UNet}~\cite{oktay2018attention},
~\textit{UNet++}~\cite{zhou2018unetpp}, and ~\textit{TransUNet}~\cite{chen2021transunet}.

\noindent\textbf{Baselines.}
We compared against: (1)~\textit{Clean};
(2)~\textit{EM}~\cite{huang2021unlearnable}; 
(3)~\textit{LSP}~\cite{yu2022availability}; (4)~\textit{SEP}~\cite{liu2023stable};
(5)~\textit{TAP}~\cite{fowl2021adversarial};
(6)~\textit{PUE}~\cite{wang2025provably}; 
(7)~\textit{UMed}~\cite{lin2024safeguarding}. All baselines were
adapted to 3D volumetric inputs.

\noindent\textbf{Metrics and Implementation.}
We reported the Dice Similarity Coefficient (DSC, \%) and 95th-percentile Hausdorff Distance (HD95, mm). Lower DSC and higher HD95 on victim models reflect more effective data protection. The perturbation budget was
$\epsilon=4/255$. We trained VoxShield for 100 epochs with Adam optimizer (lr$=$1e-4) on a single NVIDIA
RTX 5090 GPU. The loss weights were set to $\lambda_{\text{ISC}}=0.2$ and $\lambda_{\text{SPD}}=0.05$.

\subsection{Main Results}

\begin{table}[t]
\centering
\caption{Protection effectiveness on victim 3D\,UNet.
  DSC\,(\%) $\downarrow$ and HD95\,(mm) $\uparrow$ indicate stronger
  protection; PSNR\,(dB) $\uparrow$ and SSIM $\uparrow$ measure
  perturbation imperceptibility.
  Best UE results in \textbf{bold}.}\label{tab:main}
\setlength{\tabcolsep}{3pt}
\footnotesize
\begin{tabular}{l cccc cccc}
\toprule
\multirow{2}{*}{Method}
  & \multicolumn{4}{c}{BraTS19}
  & \multicolumn{4}{c}{FLARE21} \\
\cmidrule(lr){2-5}\cmidrule(lr){6-9}
  & DSC$\downarrow$ & HD95$\uparrow$
  & PSNR$\uparrow$ & SSIM$\uparrow$
  & DSC$\downarrow$ & HD95$\uparrow$
  & PSNR$\uparrow$ & SSIM$\uparrow$ \\
\midrule
Clean
  & 80.0 &  16.0 & --   & --
  & 88.6 &  10.2 & --   & -- \\ \midrule
EM~\cite{huang2021unlearnable}
  & 69.3 & 30.2 & 39.7   & 0.9219
  & 57.1 & 90.1 & 30.0   & 0.9312 \\
LSP~\cite{yu2022availability}
  & 80.5 & 15.0 & 39.6   & 0.9916
  & 88.5 & 17.5 & 43.0   & 0.9970 \\
SEP~\cite{liu2023stable}
  & 72.3 &  16.7 & 33.9   & 0.7544
  & 88.4 &  15.7 & 40.8   & 0.9511 \\
TAP~\cite{fowl2021adversarial}
  & 68.4 &  22.7 & 38.0   & 0.8792
  & 45.7 &  136.7 & 37.7   & 0.9084 \\
PUE~\cite{wang2025provably}
  & 60.0 &  26.5 & 40.2   & 0.9279
  & 36.1 &  150.2 & 37.9   & 0.9155 \\
UMed~\cite{lin2024safeguarding}
  & 40.2 &  34.7 & 54.9   & 0.9990
  & 26.3 &  107.3 & 48.0   & 0.9973 \\
\midrule
\textbf{VoxShield}
  & \textbf{$<1e^{-3}$}  & \textbf{217.4} & \textbf{56.3} & \textbf{0.9994}
  & \textbf{6.8} & \textbf{172.4}    & \textbf{49.0} & \textbf{0.9980} \\
\bottomrule
\end{tabular}
\end{table}

Table~\ref{tab:main} reported the protection performance on both datasets. VoxShield achieved a dramatic
reduction in victim model DSC, substantially outperforming all baselines
on both datasets. 2D-UE (EM, SEP, LSP, PUE, UMed) showed limited effectiveness on 3D
data, because their slice-independent noise patterns were partially smoothed by 3D convolution kernels,
confirming our hypothesis that disrupting inter-slice consistency is essential for effective 3D protection.

\begin{table}[t]
\centering
\caption{Cross-architecture transferability on FLARE21 (Mean DSC\,\%). Noise generated with UNet surrogate, evaluated on different victim architectures.}\label{tab:transfer}
\setlength{\tabcolsep}{6pt}
\begin{tabular}{lcccc}
\toprule
Method & 3D-UNet & Att-UNet & UNet++ & TransUNet \\
\midrule
Clean     & 88.6 & 83.2   & 83.4   & 80.7 \\
EM       & 57.1   & 48.6   & 68.6   & 74.2 \\
\textbf{VoxShield}  & \textbf{6.8} & \textbf{9.0} & \textbf{10.8} & \textbf{3.1} \\
\bottomrule
\end{tabular}
\end{table}
\noindent\textbf{Cross-Architecture Transferability.}
Table~\ref{tab:transfer} demonstrated that VoxShield perturbations transferred effectively across
architectures. Since the perturbations targeted the fundamental volumetric continuity prior shared by
\textit{all} 3D architectures (not architecture-specific features), the protection generalized from the
UNet surrogate to Attention UNet, UNet++, and even Transformer-based TransUNet.

\noindent\textbf{Ablation Study.}
Table~\ref{tab:ablation} validated the contribution of each component on FLARE21. Removing
$\mathcal{L}_{\text{ISC}}$ caused the largest performance drop, confirming that inter-slice spectral
diversity is the most critical attack vector against 3D segmentation. Removing $\mathcal{L}_{\text{SPD}}$
also degraded protection, as the generator lost its ability to corrupt deep semantic representations.

\begin{figure}[t]
\centering
\includegraphics[width=\textwidth]{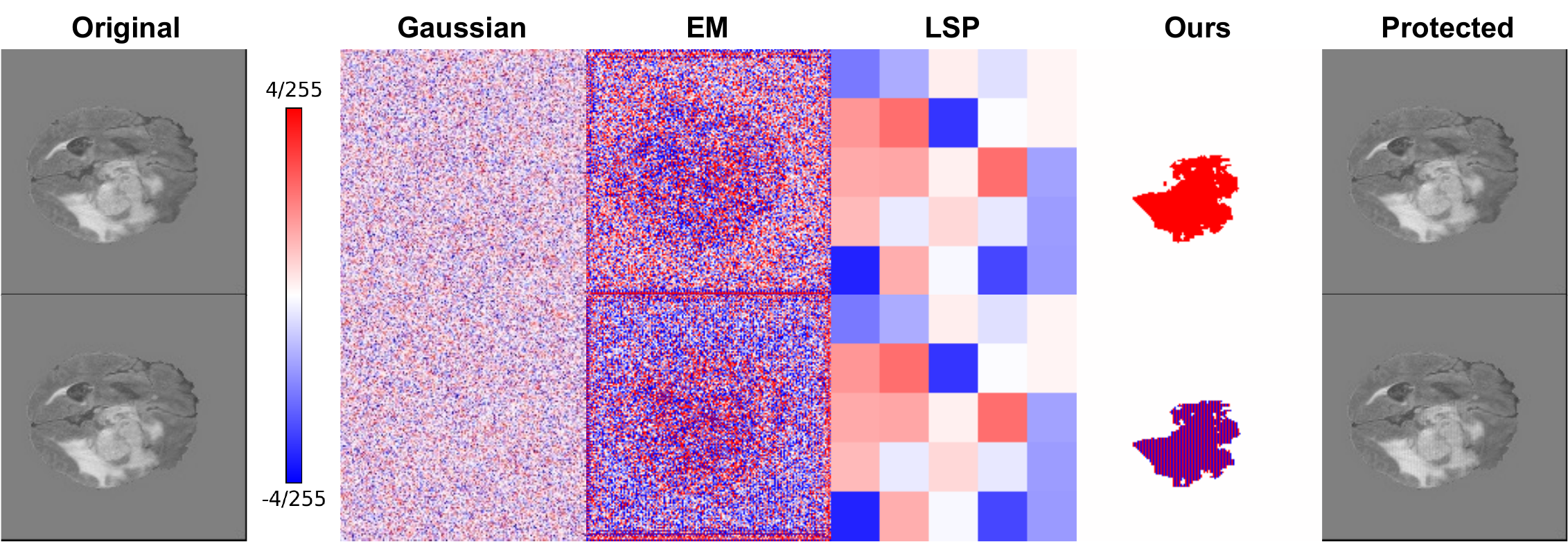}
\caption{Visualization of BraTS19 noise patterns across consecutive $z$-slices.}
\label{fig:isc}
\end{figure}

\noindent\textbf{Inter-Slice Frequency Consistency Disruption.}
We visualized the noise patterns across consecutive $z$-slices for different methods.
VoxShield produced spectrally diverse perturbations that shifted drastically between
adjacent slices, while 2D methods (EM, SEP, LSP, PUE, UMed) generated temporally coherent noise that
3D kernels could readily filter. Fig.~\ref{fig:isc} illustrates these differences.

\begin{table}[t!]
\centering
\caption{Ablation study on FLARE21. Best results in \textbf{bold}.}\label{tab:ablation}
\setlength{\tabcolsep}{6pt}
\begin{tabular}{lcc}
\toprule
Configuration & DSC\,(\%)\,$\downarrow$ & HD95\,(mm)\,$\uparrow$ \\
\midrule
\textbf{VoxShield (full)}
  & \textbf{6.8}  & \textbf{172.4} \\
\midrule
w/o $\mathcal{L}_{\text{SPD}}$ (no logits divergence)
  & 24.8    & 113.5 \\
w/o $\mathcal{L}_{\text{ISC}}$ (no ISC disruption)
  & 32.9  & 168.6 \\
\bottomrule
\end{tabular}
\end{table}

\section{Conclusion}

To the best of our knowledge, VoxShield is the first UE framework that explicitly targets inter-slice continuity priors in 3D medical segmentation. By targeting inter-slice anatomical consistency and effectively degrading segmentation performance, VoxShield creates the noise to disrupt structural continuity and semantic predictions.
VoxShield achieves the strongest protection among the evaluated baselines, demonstrating that securing 3D data requires attacking the volumetric priors fundamental to 3D models.

\begin{credits}

\subsubsection{\ackname}
This work was supported by Zhejiang Leading Innovative and Entrepreneur Team Introduction Program (2024R01007)  and the “Pioneer” and “Leading Goose” Research and Development Program of Zhejiang (2025C02077). 

\subsubsection{\discintname}
The authors have no competing interests to declare that are relevant to the content of this article.
\end{credits}

\bibliographystyle{splncs04}
\bibliography{reference}

\end{document}